\documentclass{bmvc2k}

\title{Multispectral Pedestrian Detection via Simultaneous Detection and Segmentation}

\addauthor{Chengyang Li}{licy_cs@zju.edu.cn}{1}
\addauthor{Dan Song}{songdan1992@zju.edu.cn}{1}
\addauthor{Ruofeng Tong}{trf@zju.edu.cn}{1}
\addauthor{Min Tang}{tang_m@zju.edu.cn}{1}

\addinstitution{
 State Key Lab of CAD\&CG\\
 Zhejiang University\\
 Hangzhou, China
}

\runninghead{Li et al.}{Multispectral Pedestrian Detection via SDS}

\usepackage{wrapfig}
\usepackage{multirow}

\begin{document}

\maketitle

\begin{abstract}
Multispectral pedestrian detection has attracted increasing attention from the research community due to its crucial competence for many around-the-clock applications (e.g., video surveillance and autonomous driving), especially under insufficient illumination conditions. 
We create a human baseline over the KAIST dataset and reveal that there is still a large gap between current top detectors and human performance.
To narrow this gap, we propose a network fusion architecture, which consists of a multispectral proposal network to generate pedestrian proposals, and a subsequent multispectral classification network to distinguish pedestrian instances from hard negatives.
The unified network is learned by jointly optimizing pedestrian detection and semantic segmentation tasks.
The final detections are obtained by integrating the outputs from different modalities as well as the two stages.
The approach significantly outperforms state-of-the-art methods on the KAIST dataset while remain fast.
Additionally, we contribute a sanitized version of training annotations for the KAIST dataset, and examine the effects caused by different kinds of annotation errors. Future research of this problem will benefit from the sanitized version which eliminates the interference of annotation errors.
\end{abstract}

\section{Introduction}
\label{sec:intro}
Pedestrian detection is a vigorously studied topic in the field of computer vision over the past few decades, with diversified potential applications such as video surveillance, autonomous driving and robotics. 
Nevertheless, the majority of existing detectors focus on color images only, and they probably fail to work under insufficient illumination conditions, e.g., nighttime.

Long-wavelength infrared (thermal) images provide an alternative choice to address this challenge.
Thermal cameras capture the radiated heat of objects, thus they can present clear human silhouettes even in absence of natural light, yet they lose detailed visual characteristics (e.g., color and texture) that often presented by color images.
This makes color images and thermal images complementary with each other by nature.
With the introduction of the KAIST Multispectral Pedestrian Benchmark \cite{hwang2015multispectral}, multispectral pedestrian detection has attracted increasing attention from the computer vision community \cite{wagner2016multispectral,choi2016multi,BMVC2016_liu,konig2017fully,park2018unified,li2018illumination,guan2018fusion}.
Effectively fusing multispectral data for pedestrian detection is a non-trivial task.
We create a human baseline and the results indicate that even the current state-of-the-art detectors
\cite{konig2017fully,li2018illumination,guan2018fusion}
lag behind human performance by a wide gap.
Therefore, there is still a large potential to improve the detection performance by better leveraging multispectral images.

\begin{wrapfigure}{r}{6.1cm}
\includegraphics[width=5.3cm]{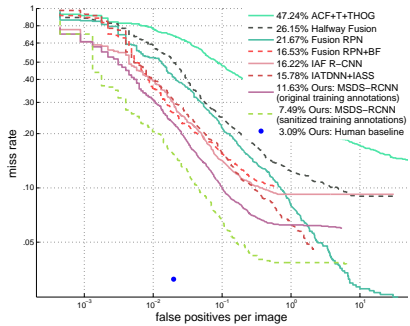}\
\caption{We have made efforts to narrow the gap between detection methods and human baseline (on the KAIST test set). Our method surpasses existing state-of-the-arts by a large margin (26\% relative). About one third of the error is attributed to the annotation noise, which can be further eliminated using our sanitized training annotations.}
\label{fig:highlight}
\end{wrapfigure}

In this work, we investigate how to effectively and efficiently detect pedestrians by leveraging RGB-thermal pairs with convolutional neural networks (convnets). 
We propose a network fusion architecture for multispectral pedestrian detection, which is denoted as the Multispectral Simultaneous Detection and Segmentation R-CNN (MSDS-RCNN).
Specifically, MSDS-RCNN consists of two multispectral fusion networks, among where the former network is responsible for generating candidate proposals and the latter network focuses on handling hard examples.
Recent work \cite{brazil2017illuminating,du2017fused,mao2017can} has shown that semantic segmentation is beneficial for RGB based pedestrian detection.
We not only confirm their conclusions on multispectral pedestrian dataset but also reveal that incorporating semantic segmentation task in proposal stage is credited for the majority of the performance improvement.

The noise of training annotation is a nonnegligible factor that could lead to performance degeneration. We manually sanitize the KAIST training annotations. Taking MSDS-RCNN as a baseline, we examine the effects of different kinds of training annotation errors, including imprecise localization, misclassification and misaligned regions.

Our major contributions are fourfold:
First, we introduce an effective and efficient architecture, called MSDS-RCNN, for multispectral pedestrian detection.
Second, we create a human baseline for the KAIST dataset to reveal the gap between current detectors and human performance.
Third, we provide a sanitized version of training annotations for the KAIST dataset and based on which, the effects of training annotation quality are evaluated.
Last but not least, our MSDS-RCNN pushes the state-of-the-art performance on KAIST dataset from 15.78\% to 11.63\% in terms of log-average miss rate (26\% relative reduction). Using the sanitized training annotations, the detection performance can be further boosted to 7.49\%.

\section{Related Work}
\label{sec:relat}
\noindent {\bf Color Image based Pedestrian Detection:}
As a canonical case of general object detection, pedestrian detection is one of the hot topics in computer vision \cite{benenson2014ten,nguyen2016human}. 
The majority of past work for pedestrian detection is based on color image and recent top performing detectors are typically variants of Fast/Faster R-CNN \cite{girshick2015fast,ren2015faster}.
MS-CNN \cite{cai2016unified} and SAF-RCNN \cite{li2018scale} handled the scale-variance problem via specifically designed multi-scale sub-networks. 
Zhang et al. \cite{zhang2016faster} showed that the under-performance of Faster R-CNN in pedestrian detection task is attributed to the Fast R-CNN classifier due to insufficient input resolution and lack of bootstrapping strategy. 
Competitive results can be achieved by cascading boosted forest \cite{appel2013quickly} on top of the high-resolution RPN feature maps, denoted as RPN+BF. 
Zhang et al. \cite{zhang2017citypersons} revealed that after proper adaptations such as pedestrian-specific RPN scales and input up-scaling, a plain Faster R-CNN gained substantial improvement and almost matched a state-of-the-art detector.
F-DNN \cite{du2017fused} and SDS-RCNN \citep{brazil2017illuminating}
used separate downstream classifiers that do not share weights with the proposal network, so that they can better handle hard examples. In this paper we will explore this insight in the scope of multispectral pedestrian detection.

\noindent {\bf Multispectral Pedestrian Detection:}
Since the release of the KAIST Multispectral Pedestrian Benchmark \cite{hwang2015multispectral}, there is a growing interest in pedestrian detection leveraging aligned color and thermal images.
The initial baseline ACF+T+THOG was extended from the aggregated channel features (ACF) \cite{dollar2014fast} and augmented with thermal channels.
Wagner et al. \cite{wagner2016multispectral} adopted ACF+T+THOG to generate region proposals, which were then re-scored by a convnet.
Choi et al. \cite{choi2016multi} first used separate RPNs to generate proposals on color and thermal images and then evaluated them with support vector regression (SVR).
A later extension \cite{park2018unified} reformulated shallow modules as network architectures so that it can be trained end-to-end.
Liu et al. \cite{BMVC2016_liu} explored different network fusion architectures developed from Faster R-CNN and discovered that halfway fusion produced best performance.
K{\"o}nig et al. \cite{konig2017fully} extended RPN+BF to multispectral pedestrian detection and proposed Fusion RPN+BF. 
Almost at the same time, \cite{li2018illumination,guan2018fusion} proposed illumination-aware fusion architectures that fused the outputs from color/thermal sub-networks or day/night sub-networks by a illumination-aware weighted function.
In this work, we do not incorporate such illumination-aware weighting mechanism, yet our detection performance already surpasses theirs remarkably. Additional improvement can be expected if we adopt such mechanism in our approach.

\noindent {\bf Segmentation for Pedestrian Detection:}
Object detection and semantic segmentation are two highly correlated tasks and recently researchers have explored utilizing semantic segmentation for pedestrian detection.
Since many pedestrian datasets do not provide segmentation masks, initial attempts \cite{costea2016semantic,zhang2017citypersons,mao2017can,du2017fused,hu2017pushing} obtained segmentation using models pretrained on segmentation datasets such as Cityscapes \cite{cordts2015cityscapes}, MS-COCO \cite{lin2014microsoft} and CamVid \cite{brostow2009semantic}, and then took the generated masks as additional cue for inference.
Recent work \citep{brazil2017illuminating} resorted bounding box annotations of pedestrians as weak segmentation mask supervision, thus segmentation and detection tasks can be simultaneously trained by optimizing a joint loss function.

\section{Preliminaries}
\label{sec:pre}

\subsection{Pedestrian Benchmark}
\label{subsec:data}
In this work we focus on the KAIST dataset \cite{hwang2015multispectral}, which contains 95,328 aligned color-thermal image pairs, with manual annotations amount to a total of 103,128 bounding boxes covering 1,182 unique pedestrians. 
Following the method presented in \cite{li2018illumination}, we sample images every 2 frames from training videos, exclude heavily occluded, truncated and small (< 50 pixels) pedestrian instances, and finally obtain 7,601 training images.
The test set contains 2,252 images sampled every 20th frame from videos, among which 1,455 images are captured during daytime and the other 797 images are during nighttime. 
For evaluation, we strictly follow the reasonable setting provided by the KAIST benchmark, and measure the log-average miss rate (MR) over the range of [$10^{-2}$, $10^{0}$] false positives per image (FPPI). 
Since the original annotations of the test set contain many problematic bounding boxes, we use the improved annotations provided by Liu et al. \cite{imp_annot} to enable a reliable comparison.

\subsection{Human Baseline}
\label{subsec:human}
Before delving into our methodology, we try to figure out how much potential is a detector expected to improve.
To this end, we construct a human baseline by asking human annotators to `detect' on the KAIST test set, which can be viewed as a perfect detector.
Considering that existing detectors are based on single image (color-thermal pair), we present frames in random order to the human annotators so that surrounding or temporal information is inaccessible.
Since pedestrian instances might be invisible in one modality, we ask human annotators to double check both color and thermal images before drawing their detections.
As expected, human performance widely surpasses existing state-of-the-arts.
At \textasciitilde 0.02 FPPI, the current top performing detectors \cite{konig2017fully,li2018illumination,guan2018fusion} produce 8$\times$ miss rate than human baseline (see Figure~\ref{fig:highlight}), indicating the automatic detector still has a large potential to improve.
The superior of human performance owes to their priori knowledge, for example, a human annotator can easily distinguish human figure sculptures from real persons.
Detection algorithms are expected to at least approach human performance.

\section{Proposed Method}
\label{sec:method}
The proposed network architecture consists of two components: a multispectral proposal network (MPN)
and a multispectral classification network (MCN). The overview of the proposed MSDS-RCNN is illustrated in Figure~\ref{fig:overview} and the details are explained below.

\subsection{Multispectral Proposal Network}
\label{subsec:mpn}
The MPN aims to generate candidate bounding boxes covering the majority of ground-truth pedestrian instances, by leveraging the information from both color and thermal modalities. Consequently, the generated proposals inevitably contain a large portion of false positives, which will be addressed by the subsequent MCN.

\begin{figure}
\centering
\includegraphics[width=10.2cm]{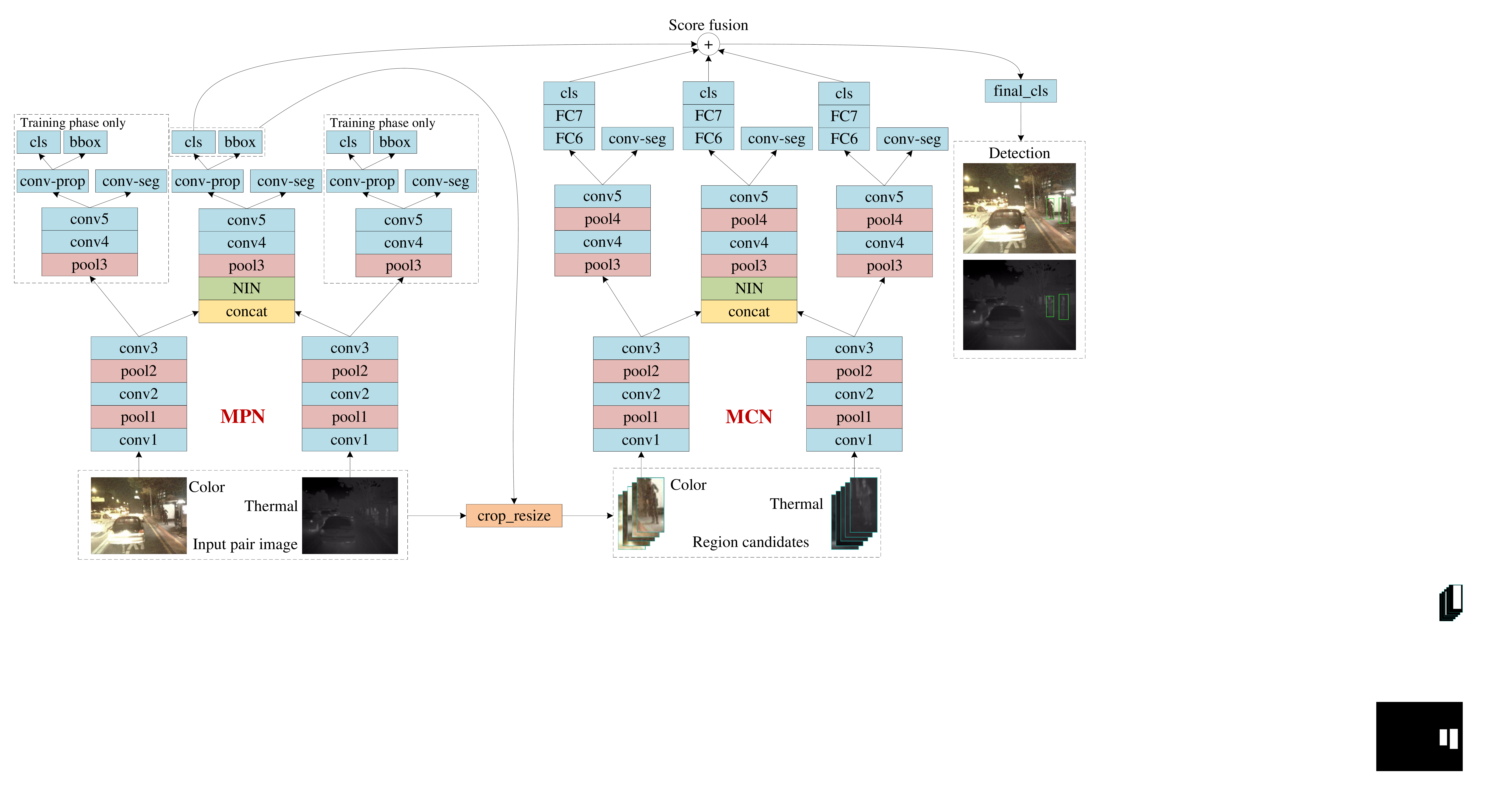}
\caption{Overview of the proposed MSDS-RCNN.}
\label{fig:overview}
\end{figure}

As shown in Figure~\ref{fig:overview}, the MPN starts from two networks separately taking the color image or thermal image as input, which are based on the VGG-16 \cite{simonyan2014very} architecture with fully connected layers removed. We fuse the two networks halfway, immediately after their third convolutional blocks, obtaining a merged stream with a balance between fine visual details and semantic information.
Network fusion is conducted by first concatenating the feature maps and then reducing dimension using Network-in-Network (NIN) \cite{lin2013network}, so that the subsequent layers in the VGG-16 architecture can be reused.
We do not truncate the original color stream and thermal stream during training phase, since they can be used to provide more diversified proposals for training the subsequent MCN.
We further remove the fourth pooling layer to provide a finer feature stride of 8, which is shown beneficial for handling small instances \cite{BMVC2016_liu,zhang2017citypersons}.
For each of color stream, thermal stream and merged stream, we build a standard proposal module on the top of each conv5\_3 layer in VGG-16 architecture, which consists of a $3 \times 3$ intermediate convolutional layer followed by two sibling
$1 \times 1$ convolutional layers for bounding box regression and classification respectively \cite{ren2015faster}.
The anchors are tailed for pedestrian detection as follows.
We split the full scale range of training data into 8 quantile bins and use the resulting 9 endpoints as RPN scales.
Besides, we use a fixed aspect ratio of 0.41 following \cite{zhang2016faster}.
An anchor is assigned a positive label if it has an Intersection-over-Union (IoU) higher than 0.5 with any ground-truth box. Otherwise, we assign it a negative label. 
Additionally, a segmentation module is also added to the top of each conv5\_3 layer, which is a single $1 \times 1$ convolutional layer.

The MPN is thus trained by minimizing the following joint loss function with nine terms: 

\begin{equation}
\begin{split}
\mathcal{L} =& \lambda_1\mathcal{L}_{MPNcls}^{color} + \lambda_2\mathcal{L}_{MPNcls}^{thermal} + \lambda_3\mathcal{L}_{MPNcls}^{merged} \\
+& \lambda_4\mathcal{L}_{MPNbbox}^{color} + \lambda_5\mathcal{L}_{MPNbbox}^{thermal} + \lambda_6\mathcal{L}_{MPNbbox}^{merged} \\
+& \lambda_7\mathcal{L}_{MPNseg}^{color} + \lambda_8\mathcal{L}_{MPNseg}^{thermal} + \lambda_9\mathcal{L}_{MPNseg}^{merged}
\end{split}
\end{equation}

where the first six components remain the same as the PPN loss defined in Faster R-CNN \cite{ren2015faster}, and the last three components are the pixel-level loss introduced by \cite{mao2017can}.
Let $G_{x,y}$, $P_{x,y}$ respectively represent the ground-truth and predicted segmentation masks, the segmentation loss is computed as: 
$\mathcal{L}_{seg} = \frac{1}{H \times W}\sum_{(x,y)}l(G_{x,y},P_{x,y})$, where H and W denote the size of the feature map and 
$l$ is the cross-entropy loss function.
In our experiments, we set all $\lambda_i=1$.
 
During inference, we only use the fusion stream to generate pedestrian candidates, as it remarkably speeds up the testing process without obvious performance degradation (see Section~\ref{subsec:ablation} for details).

\subsection{Multispectral Classification Network}
\label{subsec:mcn}
As a subsequent stage of the MPN, the MCN is designed to re-score the proposals generated by the MPN and it particularly focuses on handling hard examples.

Pedestrian candidates generated by the MPN with confidence score greater than 0.01 are passed to the MCN and those lower than 0.01 are filtered, for both training and inference phases.
Following \citep{brazil2017illuminating}, we pad each candidate proposal by 
a factor of 0.2 on all sides to incorporate contextual information and avoid partial cropping.
For each proposal, we scale it to a fixed size before taken as input for the MCN.

To construct the MCN, we start with two separate networks based on VGG-16, each takes the cropped candidate regions of color image or thermal image as input. Then we fuse the two networks halfway, as we performed in the MPN.
On top of each FC7 layer in color stream, thermal stream and merged stream, an output layer is built for binary proposal classification.
A proposal is assigned a positive label if it has an IoU higher than 0.7 with any ground-truth box.
Otherwise, we assign it a negative label. 
Also, a segmentation module is added to each conv5\_3 layer, as we did in the MPN.
The final loss for the MCN is thus computed as: 

\begin{equation}
\begin{split}
\mathcal{L} =& \lambda_1\mathcal{L}_{MCNcls}^{color} + \lambda_2\mathcal{L}_{MCNcls}^{thermal} + \lambda_3\mathcal{L}_{MCNcls}^{merged} \\
+& \lambda_4\mathcal{L}_{MCNseg}^{color} + \lambda_5\mathcal{L}_{MCNseg}^{thermal} + \lambda_6\mathcal{L}_{MCNseg}^{merged}
\end{split}
\end{equation}

where the first three components are the classification loss and the last three components are the pixel-level segmentation loss averaged on batch instances.
We set all $\lambda_i=1$ in our experiments.

For efficiency purpose, we remove the fifth pooling layer from the VGG-16 architecture, modify the filter size of the fourth pooling layer to $2 \times 1$ and then adjust the input size to $112 \times 56$. During inference, we take top $K$ proposals as input to further reduce computational cost.
We should mention that if no more than $K$ proposals generated from the MPN exist after filtering by the confidence threshold of 0.01, we take the remaining proposals as input.

Since color and thermal modalities exhibit different visual features, it is expected that the classification characteristics from color, thermal and merged streams to be complementary when fused.
Moreover, as the MPN and the MCN is designed for handling general cases and hard examples respectively, the classification results from the MPN and the MCN are also complementary.
Therefore, we fuse the classification scores from different stages and modalities.
Given the predicted 2-class scores from the three streams of MCN:
$S_{MCN}^{color}$ = $\{c_0^{c},c_1^{c}\}$, 
$S_{MCN}^{thermal}$ = $\{c_0^{t},c_1^{t}\}$, 
$S_{MCN}^{merged}$ = $\{c_0^{m},c_1^{m}\}$, 
as well as the ones from the MPN: 
$S_{MPN}$ = $\{c_0^{p},c_1^{p}\}$, 
the final classification score is obtained via the softmax function:

\begin{equation}
c_1^{f} = \frac{e^{(c_1^{p}+c_1^{c}+c_1^{t}+c_1^{m}))}}{e^{(c_0^{p}+c_0^{c}+c_0^{t}+c_0^{m}))}+e^{(c_1^{p}+c_1^{c}+c_1^{t}+c_1^{m}))}}
\end{equation}

\section{Experiments}
\label{sec:exp}

\subsection{Implementation Details}
\label{subsec:implem}
The proposed MSDS-RCNN is implemented in the Tensorflow \cite{abadi2016tensorflow} framework. 
The training process contains two main stages and we adopt the image-centric training scheme.
In the first stage, we train the MPN using SGD with a momentum of 0.9 and a weight decay of 0.0001. 
For each image, we randomly sample 120 anchors with the ratio of positive and negative ones as 1:5.
The MPN model is initialized with a VGG-16 model pretrained on the ImageNet dataset \cite{deng2009imagenet}.
We start training with a learning rate of 0.001, divide it by 10 after 4 epochs, and terminate training after 6 epochs.
In the second stage, we train the MCN using almost the same setting as the MPN. 
The MCN model is initialized with the MPN model generated in the first stage. 
For each image, we randomly sample 60 proposals generated from the MPN with the ratio of positive and negative ones as 1:2. 
During inference, we set input image scale $S = 600$ pixels for the MPN and the number of proposals $K = 50$ for the MCN, considering the speed/accuracy trade-off (see Section~\ref{subsec:ablation} for explanations). 
Since semantic segmentation masks are unavailable in the KAIST dataset, we use pedestrian bounding box annotations as weak segmentation ground-truth masks following \citep{brazil2017illuminating}. We consider the `person', `person?' and `people' categories in the KAIST dataset as foreground, and the remaining classes as background.
We report the averaged performance after repeating the experiments for 5 times.

\subsection{Comparisons with State-of-the-arts}
\label{subsec:comp}
We evaluate the proposed MSDS-RCNN on the test set of KAIST, compared with 
Halfway Fusion \cite{BMVC2016_liu}, 
ACF+T+THOG \citep{hwang2015multispectral}, 
Fusion RPN \cite{konig2017fully}, 
Fusion RPN+BF \cite{konig2017fully}, 
IAF R-CNN \cite{li2018illumination} 
and IATDNN+IASS \cite{guan2018fusion}.
We also implement two single modality baselines for comparison, denoted as Color and Thermal. For implementation, we simply remove layers in the MSDS-RCNN model and corresponding components in the loss function that involve the other modality, then train a single-modality model using the identical procedure.

\begin{figure}
\centering
\begin{tabular}{ccc}
\includegraphics[width=3.9cm]{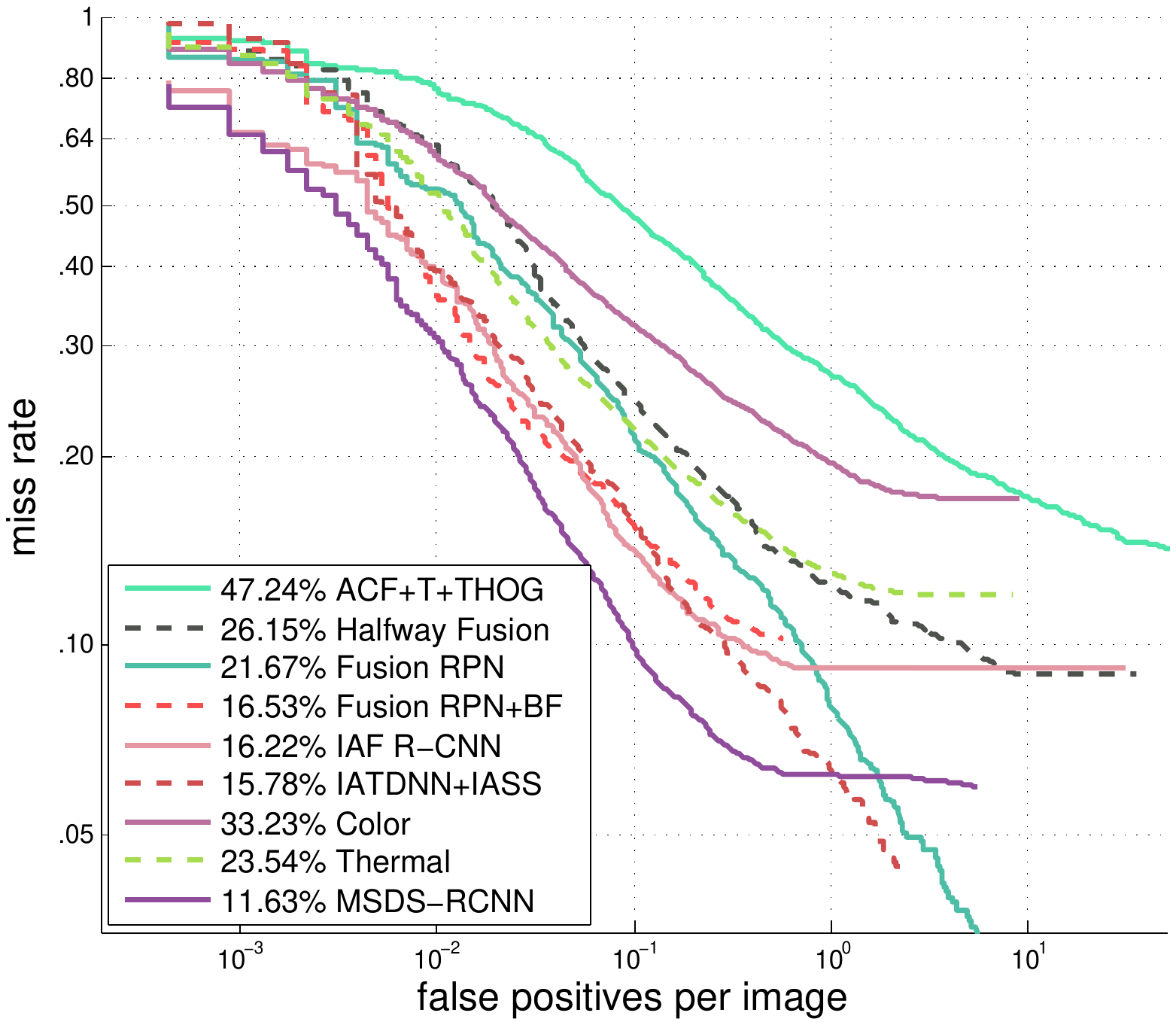}&
\includegraphics[width=3.9cm]{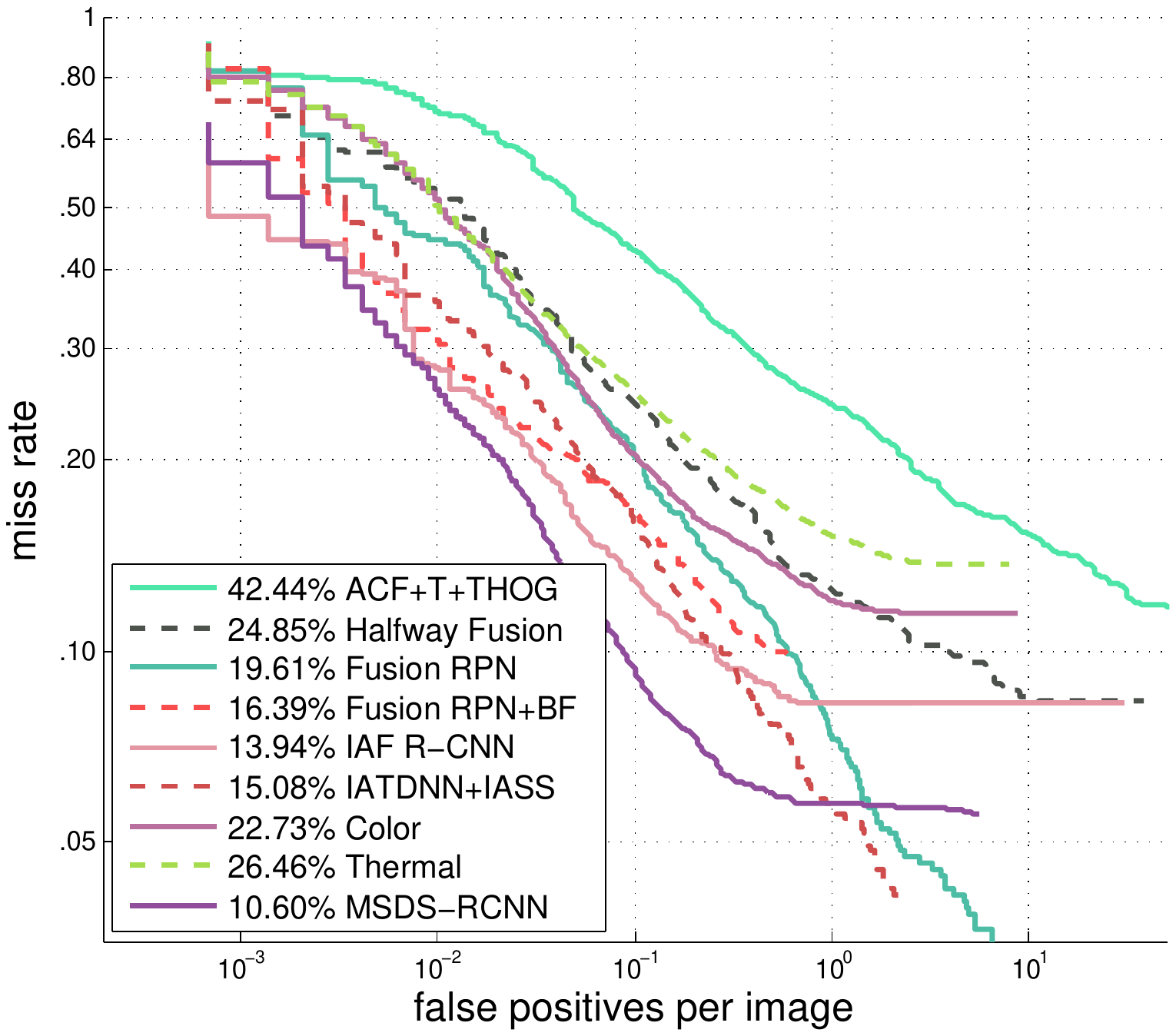}&
\includegraphics[width=3.9cm]{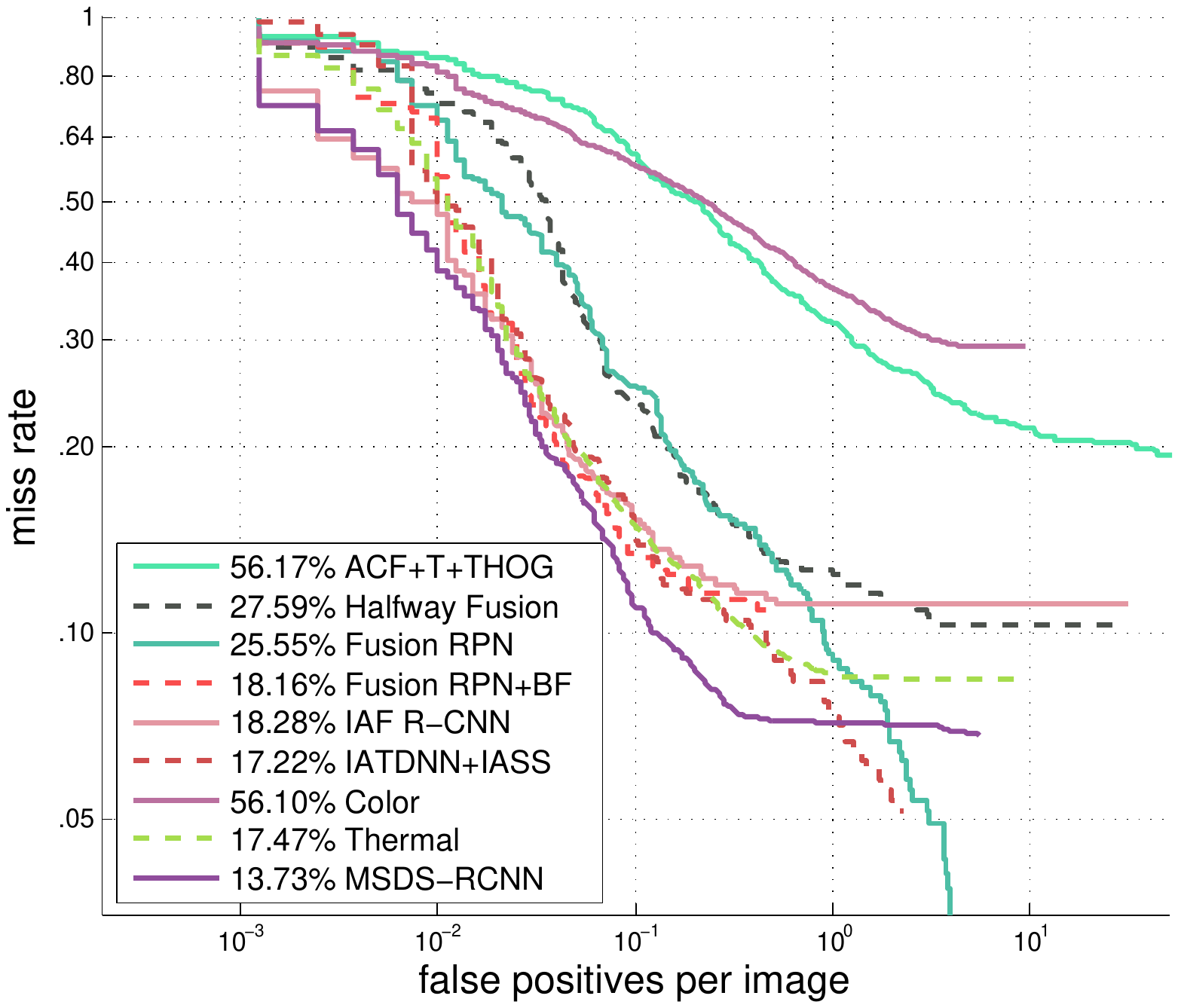}
\end{tabular}
\caption{Comparisons of detection results reported on the test set of KAIST dataset, in terms of Reasonable-all (left), Reasonable-day (middle), and Reasonable-night (right).}
\label{fig:compall}
\end{figure}

Figure~\ref{fig:compall} compares the experimental results, in terms of MR under reasonable setting. 
It can be observed that MSDS-RCNN outperforms all existing methods and single-modality baselines by a large margin, both on daytime images and nighttime images.
IATDNN+IASS is the best among existing detectors, with 15.78\% MR.
With the proposed method, we obtain 11.63\% MR, improving the current state-of-the-art by 26\% relative reduction of the error. Moreover, the efficiency of our method also surpasses IATDNN+IASS, with 228 ms/image vs. 250 ms/image on runtime with a single NVIDIA Geforce Titan X GPU.

\subsection{Ablation Studies}
\label{subsec:ablation}
This subsection is devoted to investigating the effectiveness of different design choices.

\noindent {\bf Effect of semantic segmentation:}
Table~\ref{tab:seg} compares the performance of enabling or disabling the segmentation supervision in the networks. The baseline that does not use the segmentation supervision obtains 13.59\% MR. 
By introducing the segmentation supervision in both the MPN and the MCN, the detection performance improves to 11.63\% MR, with 14\% relative reduction of the error, indicating that the segmentation supervision is also beneficial for multispectral pedestrian detection. 
We also compare the effect of introducing the segmentation supervision in the MPN or the MCN. In this case, we enable the segmentation supervision in one network and disable the other. It can be observed that in both cases we obtain performance improvement, but infusing segmentation in the MPN brings considerably greater impact than that in the MCN (12.00\% vs. 13.03\%).
We suppose this can be attributed to the coarse bounding box annotations in the KAIST dataset, which can cause more inconsistence when handling segmentation masks locally in the MCN.

\begin{table}[h]
\centering
\begin{tabular}{|c|c||c|c|c|}
\hline
\multicolumn{2}{|c||}{Supervision} & \multicolumn{3}{|c|}{Detection performance (MR)} \\
\hline
MPN & MCN & Reasonable-all & Reasonable-day & Reasonable-night \\
\hline
&& 13.59\% & 11.95\% & 16.96\% \\
$\surd$ && 12.00\% & 11.10\% & 14.14\% \\
& $\surd$ & 13.03\% & 11.74\% & 15.65\% \\
$\surd$ & $\surd$ & 11.63\% & 10.60\% & 13.73\% \\
\hline
\end{tabular}
\caption{\label{tab:seg}Effectiveness of the segmentation supervision.}
\end{table}

\noindent {\bf Effect of score fusion:}
As illustrated in Table~\ref{tab:fusion}, combining the merged stream with color stream and thermal stream pushes the performance
from 14.02\% MR to 11.95\% MR. This phenomenon reveals that although the merged stream makes use of color and thermal information, the classification characteristics of color stream and thermal stream are still complementary to the merged stream.
The scores from the MPN is also complementary, combining it slightly boosts the detection performance to 11.63\%.

\begin{table}[h]
\centering
\small
\begin{tabular}{|c|c|c|c||c|c|c|}
\hline
\multirow{2}{*}{MPN} & \multicolumn{3}{|c||}{MPN} & \multicolumn{3}{|c|}{Detection performance (MR)} \\ \cline{2-7}
& Color & Thermal & Merged & Reasonable-all & Reasonable-day & Reasonable-night \\
\hline
$\surd$ &&&& 18.88\% & 16.63\% & 22.89\% \\
& $\surd$ &&& 24.32\% & 18.43\% & 36.28\% \\
&& $\surd$ && 22.50\% & 24.56\% & 17.76\% \\
&&& $\surd$ & 14.02\% & 13.78\% & 14.54\% \\
& $\surd$ & $\surd$ & $\surd$ & 11.95\% & 10.99\% & 13.83\% \\
$\surd$ & $\surd$ & $\surd$ & $\surd$ & 11.63\% & 10.60\% & 13.73\% \\
\hline
\end{tabular}
\caption{\label{tab:fusion}Effectiveness of the score fusion scheme.}
\end{table}

\begin{figure}[htb]
\centering
\begin{tabular}{cc}
\includegraphics[width=3.9cm]{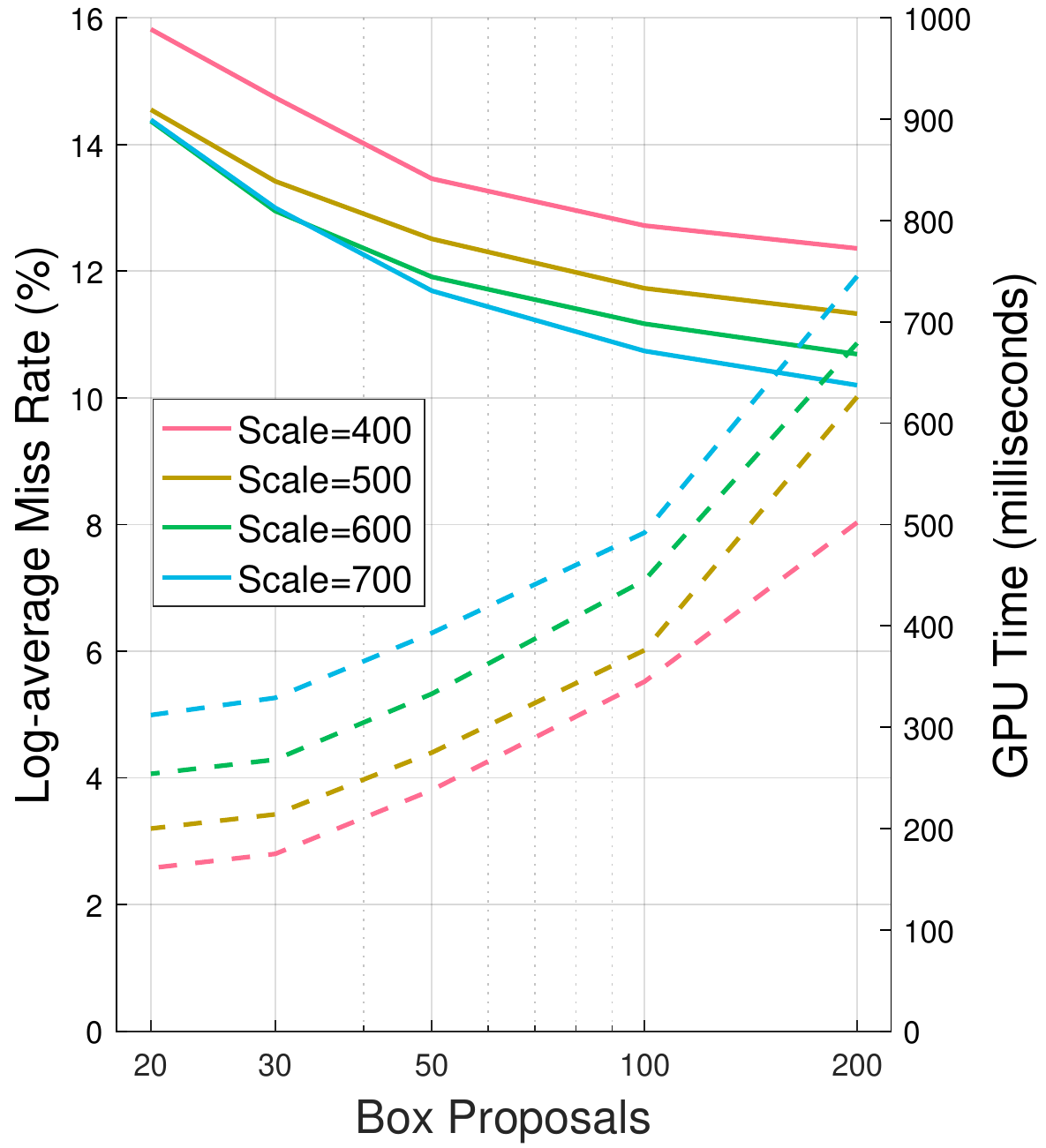}&
\includegraphics[width=3.9cm]{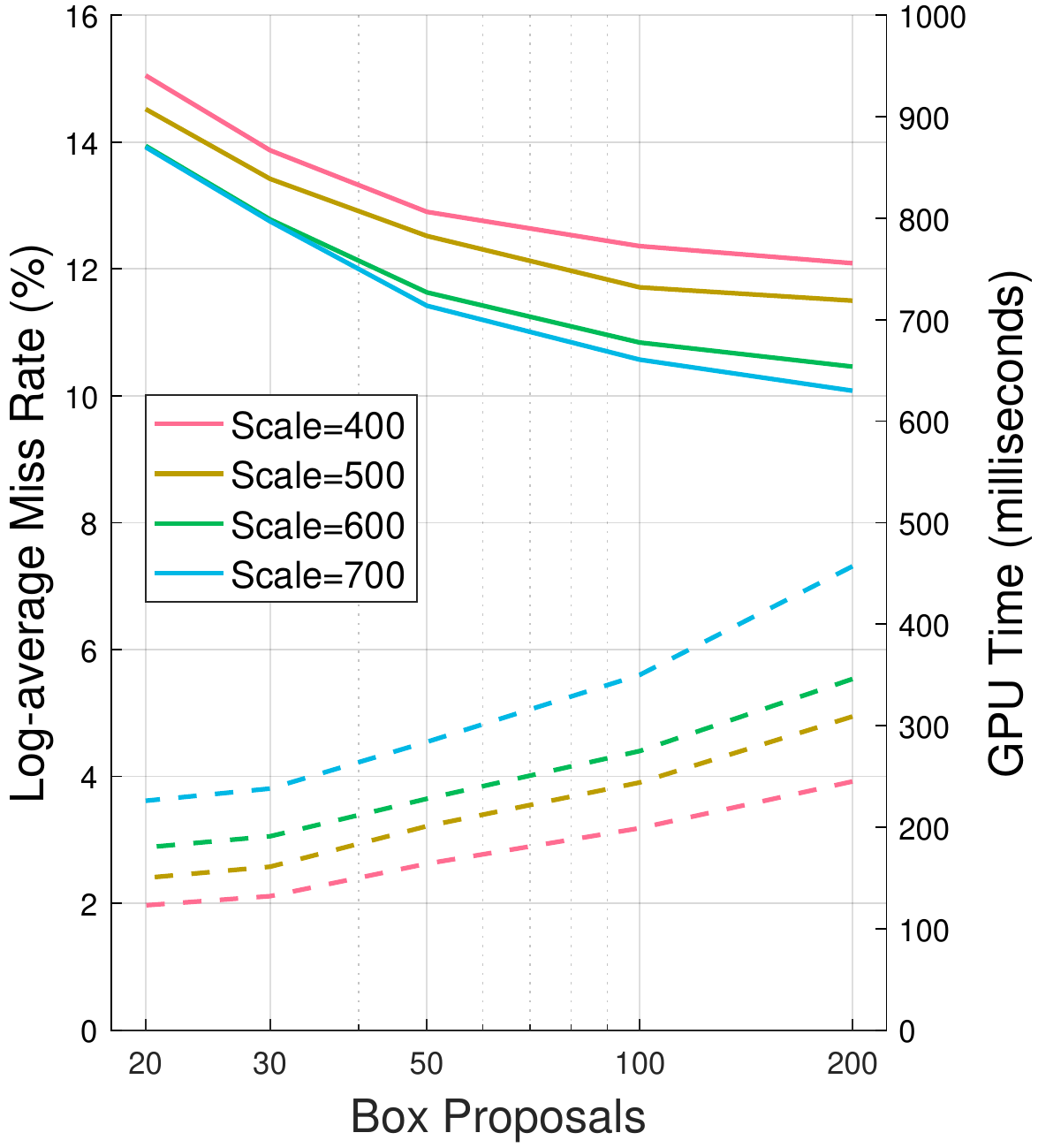}
\end{tabular}
\caption{Effect of MPN input scale, number of regions and proposal mode (left: All, right: Merged) on log-average miss rate (solid lines) and GPU inference time (dotted).}
\label{fig:trade-off}
\end{figure}

\noindent {\bf Speed/accuracy trade-off:}
Finally we evaluate the efficiency of the method. 
The runtime of the MPN is varied by the scale of input image, while that of the MCN depends on the number of input proposals.
Figure~\ref{fig:trade-off} compares the performance using different input scales and numbers of proposals.
We also compare the effects of using only the merged stream (denoted as `Merged') and using all three streams (denoted as `All') in the MPN to generate proposals.
Generally, the larger input scale or more proposals bring performance improvement but add more computational cost.
Besides, using `Merge' proposals typically obtains comparable or even better performance than `All' proposals.
Considering the speed/accuracy trade-off, we adopt 600 input image scale for the MPN, `Merge' proposal mode and 50 proposals for the MCN, which results in a process speed of 228 ms/image.

\subsection{Impact of Training Annotation Noise}
\label{subsec:noise}
The annotation noise in the KAIST dataset is a vital factor that could affect the detection performance.
The original annotations of KAIST dataset contain many problematic bounding boxes, such as missing annotation and incorrect labeling.
Hence, Liu et al. \cite{imp_annot} provided improved annotations of KAIST test set to enable a reliable evaluation.
As revealed by our previous technical report \citep{li2018illumination},
there is a big difference (>15\%) in regard of MR value between using original and improved testing annotations.
Similarly, the annotation noise in the training data would lead to error-prone optimizing process.

To study the effects of annotation noise, we create a sanitized version of KAIST training annotations.
Since annotating the whole training data is time-consuming (95,328 frames), we first filter the training images using the original annotations and obtain 7,601 valid frames (same protocol as described in Section~\ref{subsec:data}). 
Then we carefully re-label all these 7,601 frames to provide a high quality version of ground-truth annotations.
The annotation errors we corrected can be divided into three categories as follows:

\begin{figure}[b]
\centering
\begin{tabular}{ccc}
\includegraphics[width=2.8cm]{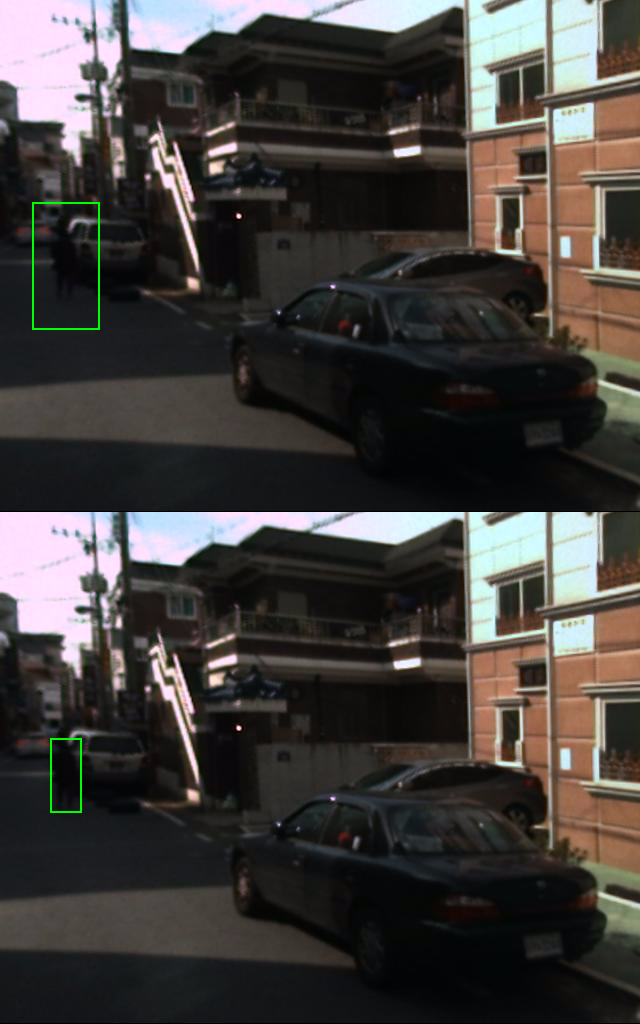}&
\includegraphics[width=2.8cm]{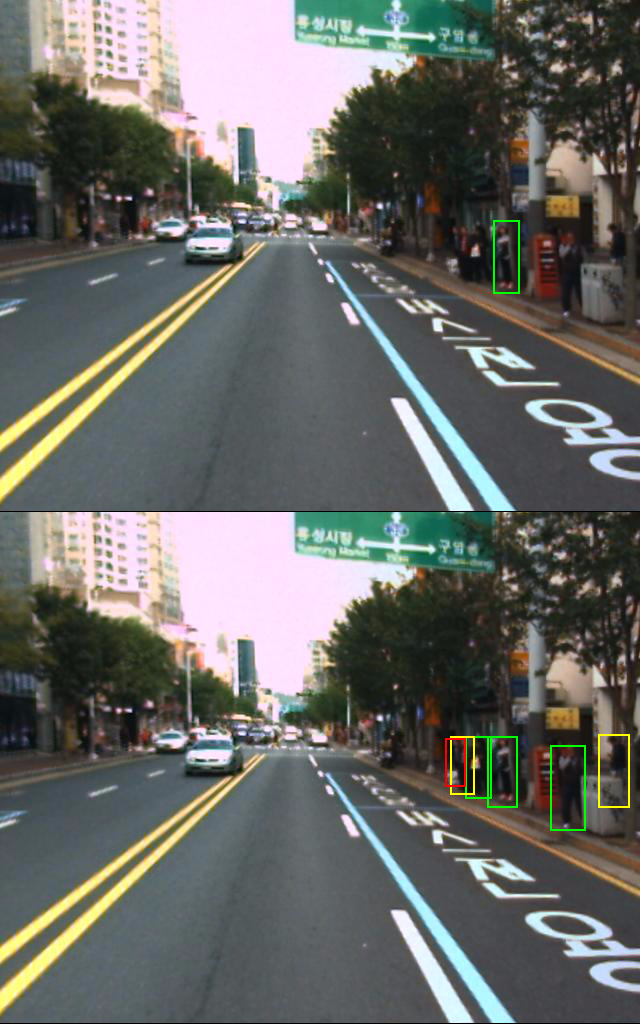}&
\includegraphics[width=5.6cm]{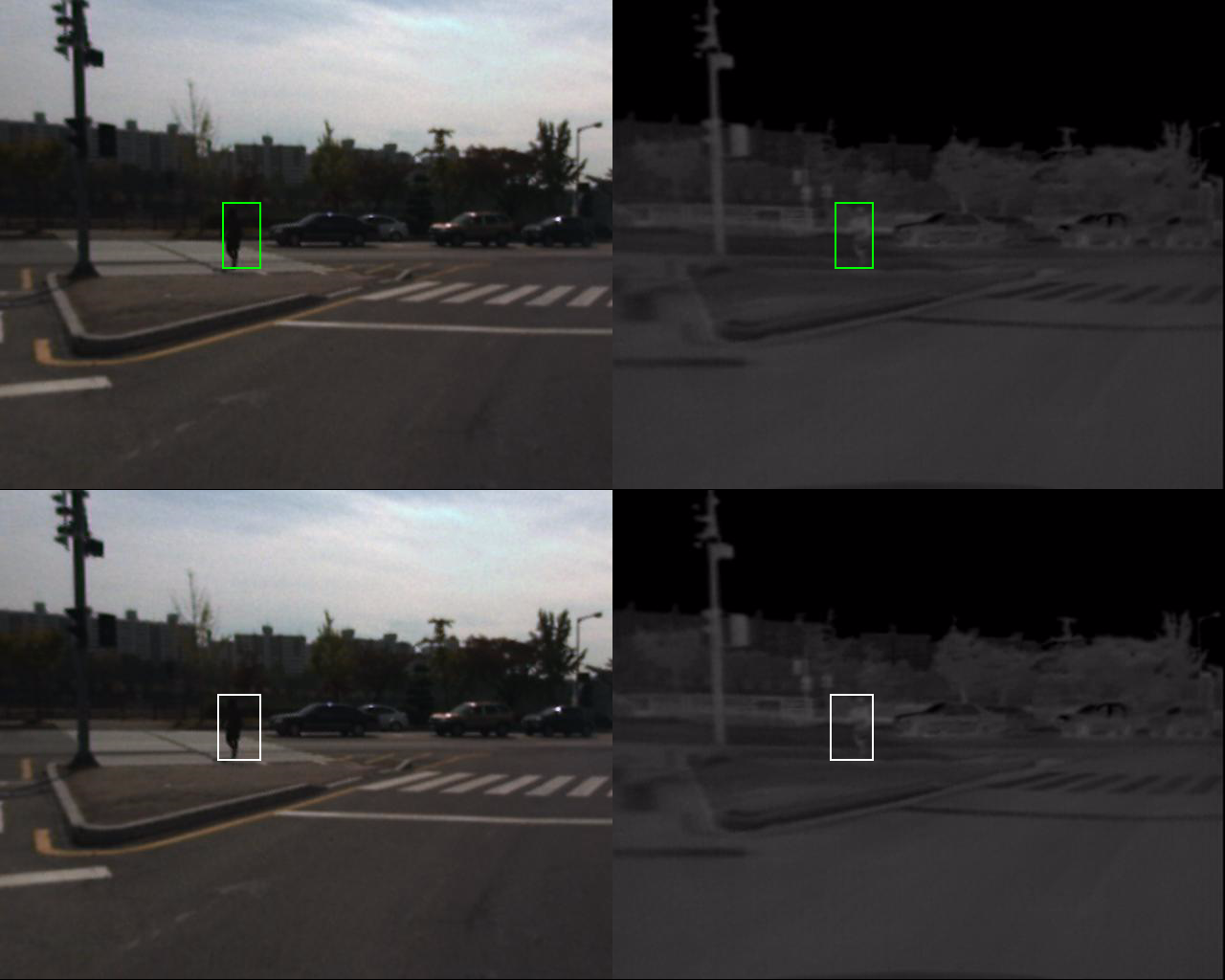}
\end{tabular}
\caption{Examples of the corrected annotations. 
Green/yellow/red bounding boxes denote non/partial/heavy-occluded pedestrians and white boxes denote ignore regions.
Top row: the original annotations. Bottom row: the sanitized annotations.}
\label{fig:examples}
\end{figure}

{\bf 1) Imprecise localization:}
In the original annotations, there are many annotated bounding boxes that do not well match the real regions of person instances.
The most common case is using a obviously larger box to annotate a small instance. 
We correct this kind of error so that each instance is tightly bounded by a box (see Figure~\ref{fig:examples} (left)).

{\bf 2) Misclassification:}
The corrected misclassification cases are those assigning an incorrect category or occlusion state to an person.
The cases also include missing annotations, i.e., incorrectly labeling a person as background (see Figure~\ref{fig:examples} (middle)).

{\bf 3) Misaligned regions:}
Although efforts have been made to ensure the paired color and thermal images align both spatially and temporally, we find there still exist cases that the multispectral images are not well aligned, particularly when the car is making a turn.
For such case, we separately label the bounding boxes in the color image and the thermal image, and then check their IoU value. 
If they have an IoU lower than 0.5, we use the minimum box that bounds both boxes to represent the instance and label it as `{\em person?a}' so that it can be ignored during training (see Figure~\ref{fig:examples} (right)).

\begin{figure}
\centering
\begin{tabular}{ccc}
\includegraphics[width=3.9cm]{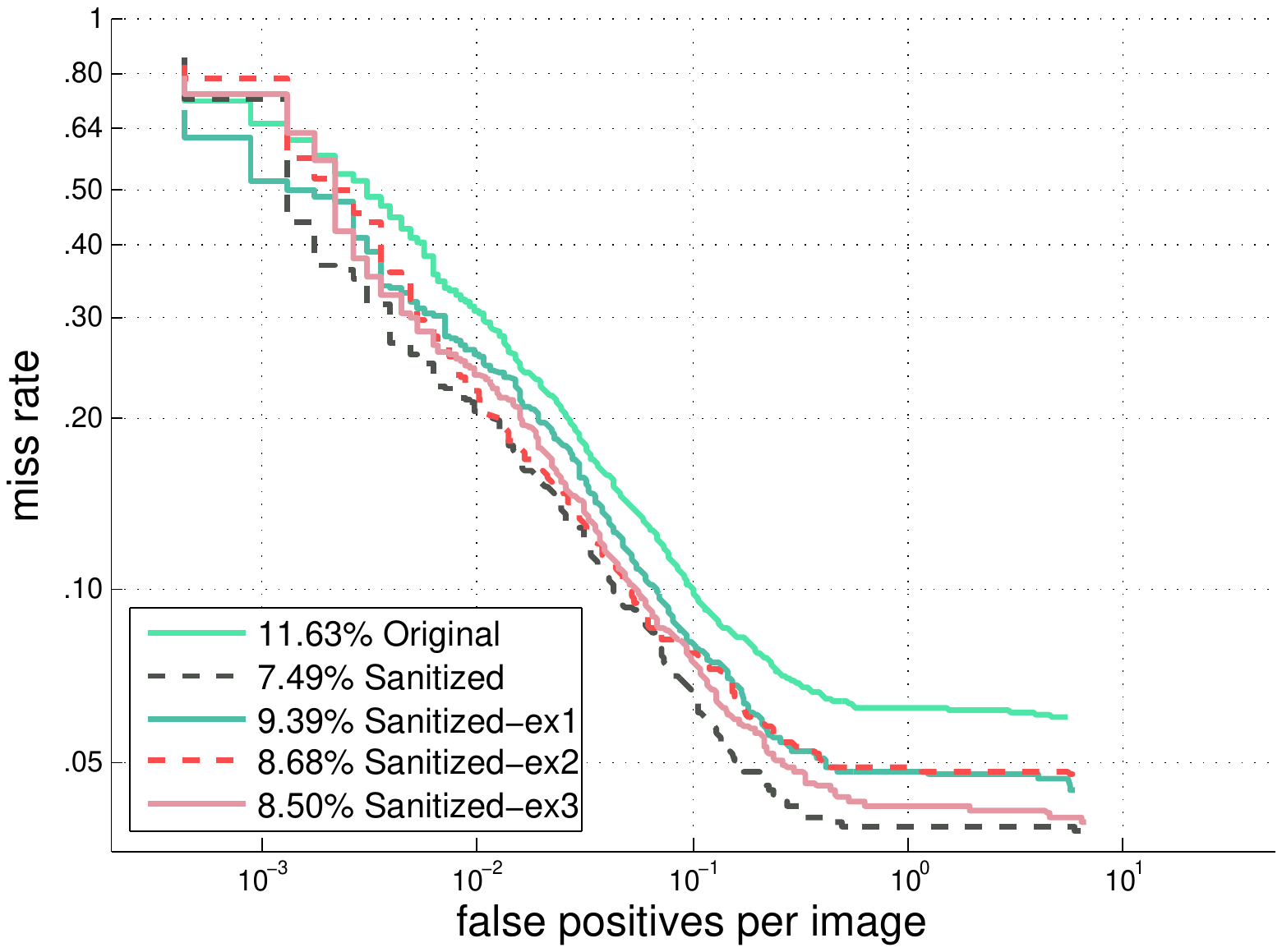}&
\includegraphics[width=3.9cm]{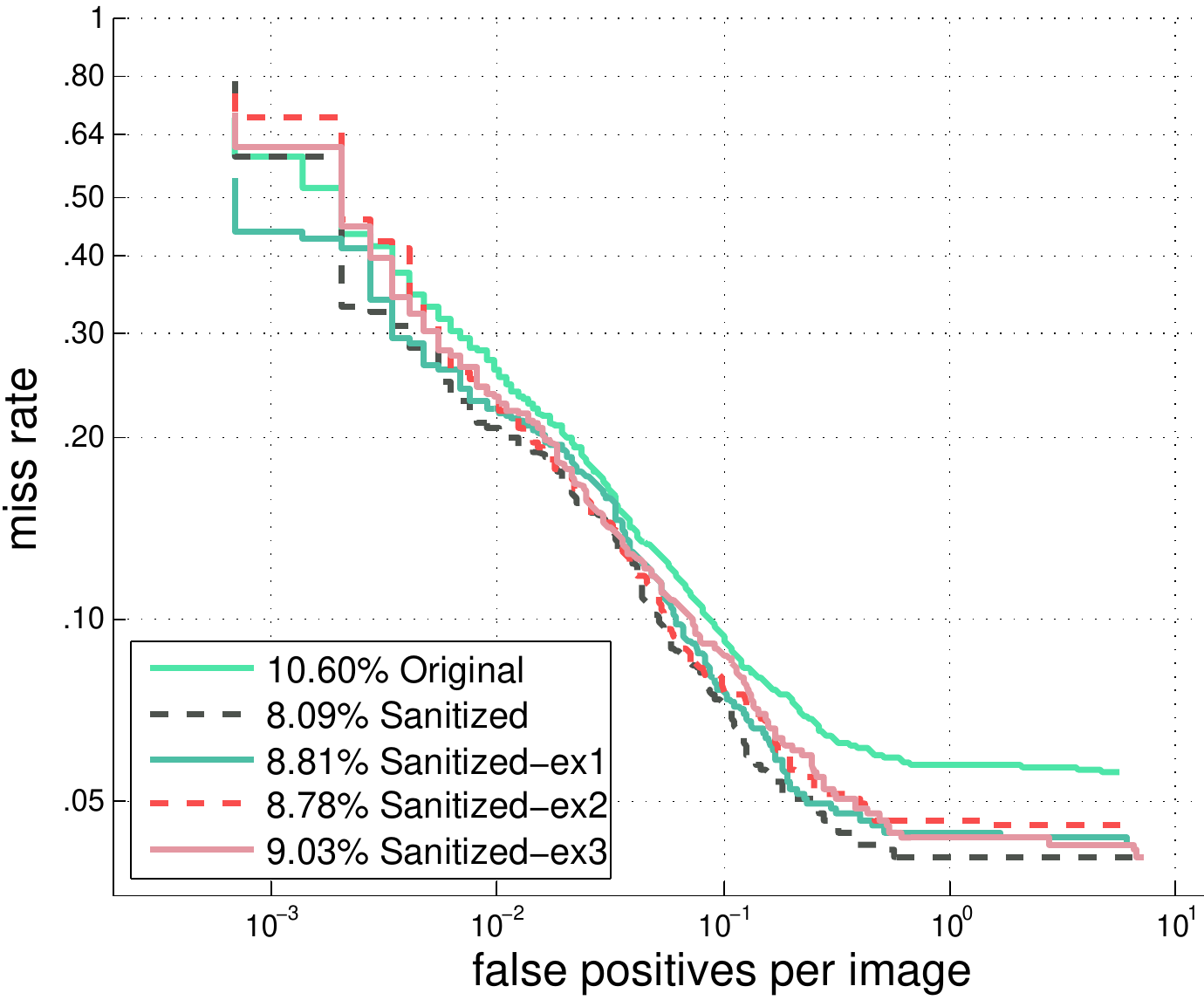}&
\includegraphics[width=3.9cm]{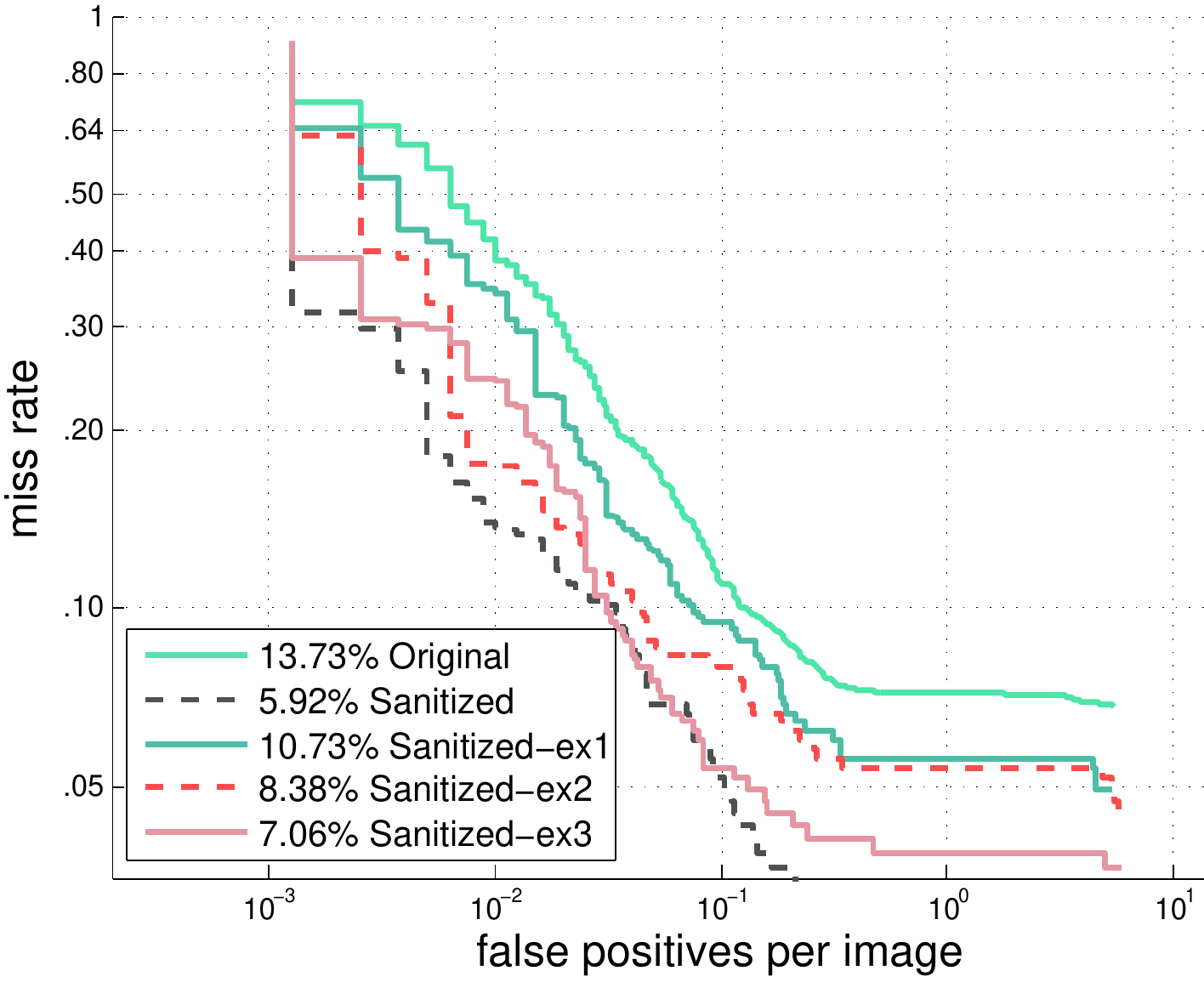}
\end{tabular}
\caption{Impact of different training annotations, compared on the test set of KAIST dataset, in terms of Reasonable-all (left), Reasonable-day (middle), and Reasonable-night (right).}
\label{fig:sanitized}
\end{figure}

Taking MSDS-RCNN as a baseline, we quantitatively study the effects of training annotation noise.
Apart from comparing the performance using the original and the sanitized annotations, we also examine the results using semi-sanitized annotations that excludes a specific type of error correction (denoted as `Sanitized-ex1', `Sanitized-ex2' and `Sanitized-ex3' respectively).
The experimental results are shown in Figure~\ref{fig:sanitized}.
Not surprisingly, using the sanitized training annotations, the detection performance improves significantly from 11.63\% MR to 7.45\% MR, which indicates that the training annotation noise is responsible for about one third of the inference error. 
For daytime images, using sanitized annotations gains 24\% relative error reduction, and the three types of annotation errors contribute quite similar degrees.
For nighttime images, using sanitized annotations obtains an amazing 56\% relative error reduction, the most of which is due to the correction of imprecise localization.

\section{Conclusion}
\label{sec:conclude}
In this work, we make efforts to narrow the gap between automatic pedestrian detectors and human performance.
We present a unified convnet fusion architecture, denoted the MSDS-RCNN, for person detection in multispectral data (color-thermal image pairs).
We show that jointly optimizing segmentation and detection tasks as well as effectively fusing the outputs from different branches
bring substantial performance improvement, leading to 26\% relative reduction of MR compared with existing state-of-the-art detector while remaining faster.
Since the original training data contains many problematic annotations, we further study the impact of training annotation noise by carefully creating a sanitized version of ground-truth annotations.
We find that the sanitized training annotations benefit the detection performance remarkably, especially for the nighttime images.
We hope that future research can benefit from the provided data.

\section*{Acknowledgement}
The research is supported in part by NSFC (61572424) and the Science and Technology Department of Zhejiang Province (2018C01080).

\bibliography{egbib}
\end{document}